# Detection and Demarcation of Tumor using Vector Quantization in MRI images

Dr. H. B. Kekre
Senior Professor,
Mukesh Patel School of Technology Management and Engineering,
SVKM's NMIIMS University
Mumbai-56, India.
hbkekre@yahoo.com

Ms. Tanuja K. Sarode
Ph.D. Scholar, MPSTME, SVKM's NMIMS University, Mumbai-56, India,
Assistant Professor, T.S.E.C.,
Mumbai-50, India.
tanuja_0123@yahoo.com

Ms. Saylee M. Gharge
Ph.D. Scholar, MPSTME, SVKM's NMIMS University, Mumbai-56, India,
Lecturer, V.E.S.I.T,
Mumbai-71 India.
sayleegarge73@yahoo.co.in

**ABSTRACT**

Segmenting a MRI images into homogeneous texture regions representing disparate tissue types is often a useful preprocessing step in the computer-assisted detection of breast cancer. That is why we proposed new algorithm to detect cancer in mammogram breast cancer images. In this paper we proposed segmentation using vector quantization technique. Here we used Linde Buzo-Gray algorithm (LBG) for segmentation of MRI images. Initially a codebook of size 128 was generated for MRI images. These code vectors were further clustered in 8 clusters using same LBG algorithm. These 8 images were displayed as a result. This approach does not leads to over segmentation or under segmentation. For the comparison purpose we displayed results of watershed segmentation and Entropy using Gray Level Co-occurrence Matrix along with this method.

## 1. INTRODUCTION

**Magnetic resonance imaging (MRI)** is primarily a medical imaging technique most commonly used in Radiology to visualize the structure and function of the body. It provides detailed images of the body in any plane. MRI provides much greater contrast between the different soft tissues of the body than does computer tomography (CT), making it especially useful in neurological (brain), musculoskeletal, and oncological (cancer) imaging. Unlike CT it uses no ionizing radiation, but uses a powerful magnetic field to align the nuclear magnetization of (usually) hydrogen atoms in water in the body. Radiofrequency fields are used to systematically alter the alignment of this magnetization, causing the hydrogen nuclei to produce a rotating magnetic field detectable by the scanner. This signal can be manipulated by additional magnetic fields to build up enough information to reconstruct an image of the body.The advantages of magnetic resonance imaging (MRI) over other diagnostic imaging modalities are its high spatial resolution and excellent discrimination of soft tissues. MRI provides rich information about anatomical structure, enabling quantitative pathological or clinical studies [1]; the derivation of computerized anatomical atlases [2]; as well as pre and intra-operative guidance for therapeutic intervention [3, 4]. Such information is also valuable as an anatomical reference for functional modalities, such as PET [5], SPECT, and functional MRI [6]. Advanced applications that use the morphologic contents of MRI frequently require segmentation of the imaged volume into tissue types. This problem has received considerable attention. Such tissue segmentation is often achieved by applying statistical classification methods to the signal intensities [7, 8]. In the ideal case, differentiation between white and gray matter in the brain should be easy since these tissue types exhibit distinct signal intensities. In practice, spatial intensity inhomogeneities are often of sufficient magnitude to cause the distributions of signal intensities associated with these tissue classes to overlap significantly. In addition, the operating conditions and status of the MR equipment frequently affect the observed intensities, causing significant inter-scan intensity inhomogeneities that often necessitate manual training on a per-scan basis. While reported methods [9, 10, 11, 12, 13, 14] have had some success in correcting intra-scan inhomogeneities, such methods require supervision for the individual scan.

The best approach to image segmentation may vary between different applications. The choice between manual, semiautomatic or fully automatic methods depends on the quality of the images, the number of objects needs to be segmented, the amount of available user time, and the required accuracy of the segmentation. The segmentation process is usually based on gray level intensity, color, shape or texture. Texture can be characterized by local variations of pixel values that repeat in a regular or random pattern on the object or image. It can also be defined as a repetitive arrangement of patterns over a region. A wide variety of texture segmentation techniques have been reported in the literature [15,16, 17-23,24]. We decided to choose a set of existing texture features [25-28] which can provide us good discriminating power and are easy to compute as compare to GLCM [29].

The work we have done is to propose a segmentation process which identifies on a MRI the opaque areas, suspect or not, present in the image using vector quantization which consumes moderate time but provide good accuracy with less complexity. Watershed algorithm has a drawback of over-segmenting the image making it obscure for identification of tumor. Segmentation using gray level co-occurrence matrix required huge time for tumor demarcation with less accuracy.

## 1.1 Vector Quantization

Vector Quantization (VQ) [30-38] is an efficient technique for data compression and has been successfully used in various applications such as index compression [39, 40]. VQ has been very popular in a variety of research fields such as speech recognition and face detection [41, 42]. VQ is also used in real time applications such as real time video-based event detection [43] and anomaly intrusion detection systems [44], image segmentation [45-48], speech data compression [49], content based image retrieval CBIR [50] and face recognition [51].

Vector Quantization (VQ) techniques employ the process of clustering. Various VQ algorithms differ from one another on the basis of the approach employed for cluster formations. VQ is a technique in which a codebook is generated for each image. A codebook is a representation of the entire image containing a definite pixel pattern which is computed according to a specific VQ algorithm. The image is divided into fixed sized blocks that form the training vector. The generation of the training vector is the first step to cluster formation on these training vectors clustering methods is applied and codebook is generated. The method most commonly used to generate codebook is the Linde-Buzo-Gray (LBG) algorithm which is also called as Generalized Lloyd Algorithm (GLA).

The rest of the paper is organized as follows. Section II describes Gray Level Co-occurrence Matrix(GLCM), Watershed algorithm and Linde Buzo Gray algorithm (LBG) algorithm used for image segmentation of MRI images. Followed by the experimental results for MRI images for comparison in section III and section IV concludes the work.

## 2. ALGORITHMS FOR SEGMENTATION

In this section we explain segmentation by Gray level co-occurrence matrix [29], basic watershed algorithm [52-56] and Linde Buzo Gray algorithm (LBG) which are used for comparative performance of tumor detection.

### 2.1 Gray Level Co-occurrence Matrix

Haralick suggested the use of gray level co-occurrence matrices (GLCM) for definition of textural features. The values of the co-occurrence matrix elements present relative frequencies with which two neighboring pixels separated by distance *d* appear on the image, where one of them has gray level *i* and other *j*. Such matrix is symmetric and also a function of the angular relationship between two neighboring pixels. The co-occurrences matrix can be calculated on the whole image, but by calculating it in a small window which scanning the image, the co-occurrence matrix can be associated with each pixel. By using gray level co-occurrence matrix we can extract different features like probability, entropy, energy, variance, inverse moment difference etc. Using co-occurrence matrix the major textural features are defined as:

Maximum Probability: $\max(P_{ij})$ (2.1)

Variance:

$$(\sum (i-\mu_i)^2 \sum P_{ij})(\sum (j-\mu_j)^2 \sum P_{ij}) \quad (2.2)$$

Correlation:

$$\sum_i \sum_j (i-\mu_x)(j-\mu_y)P_{ij} / \sigma_x \sigma_y \quad (2.3)$$

where $\mu_x$ and $\mu_y$ are means and $\sigma_x$, $\sigma_y$ are standard deviation

Entropy: $\sum_i \sum_j P_{ij} \log(P_{ij})$ (2.4)

Amongst all these features entropy has given us the best results. Hence in this paper we extracted entropy using gray level co-occurrence matrix and the results are displayed in Fig.6a alongwith that of watershed and LBG algorithms for comparison.

### 2.2 Watershed algorithm

Watershed segmentation [52] classifies pixels into regions using gradient descent on image features and analysis of weak points along region boundaries. The image feature space is treated, using a suitable mapping, as a topological surface where higher values indicate the presence of boundaries in the original image data. It uses analogy with water gradually filling low lying landscape basins. The size of the basins grows with increasing amount of water until they spill into one another. Small basins (regions) gradually merge together into larger basins. Regions are formed by using local geometric structure to associate the image domain features with local extremes measurement. Watershed techniques produce a hierarchy of segmentations, thus the resulting segmentation has to be selected using either some apriory knowledge or manually. These methods are well suited for different measurements fusion and they are less sensitive to user defined thresholds. We implemented watershed algorithm for MRI images as mentioned in [56].Results for MRI images are displayed in Fig 6(b).

## 1.2 Linde Buzo and Gray (LBG)[29, 30]

For the purpose of explaining this algorithm, we are considering two dimensional vector space as shown in Fig.1. This is obtained by considering two consecutive pixel values as x and y co-ordinates so that each pair is represented by point in x,y plane. In this algorithm centroid is computed as the first codevector $C_1$ for the training set. In Fig. 1 two vectors $v_1$ & $v_2$ are generated by adding constant error to the codevector. Euclidean distances of all the training vectors are computed with vectors $v_1$ & $v_2$ and two clusters are formed based on nearest of $v_1$ or $v_2$. Procedure is repeated for these two clusters to generate four new clusters. This procedure is repeated for every new cluster until the required size of codebook is reached or specified MSE is reached.

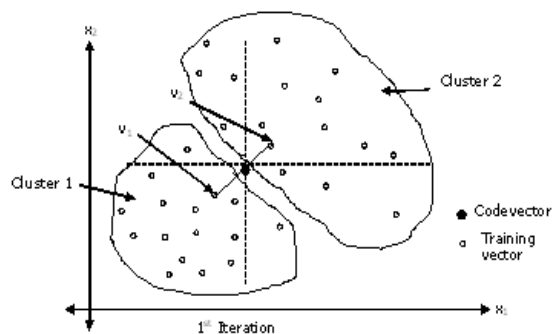

Fig.1. LBG for 2 dimensional case

In this paper initially we have selected 128 as codebook size using 12 dimensional vector space. Thus the image is divided into 128 clusters which were further reduced to 8 by using requantization. The 8 clusters thus obtained were mapped onto the image generating 8 different images representing them. On all these images Canny's operator was used to obtain the edge maps. These edge maps were superimposed on the original image giving clear demarcation of the tumor. The very first cluster gives the best results. However the other clusters also give comparatively better results as compare to watershed and GLCM algorithm.

### 3. RESULTS

Defects in the metabolic system can lead to waste build-up that can cause altered levels of consciousness (ALC). Drug exposure is a common cause for ALC. Drug-induced ALC can result from an overdose of either over-the-counter or illegal drugs. Alcohol intoxication is probably the most common cause of drug-induced ALC. Structural abnormalities of the brain can lead to ALC [Figure 2a]. Tumors (benign or malignant) can form and crowd out the normal structures of the brain. As a result, weakness in the walls of the blood vessels in the brain (aneurysms) may begin to swell, or may even break, causing blood to pool inside the head and push the brain against the bony wall of the skull. The resulting damage can then cause ALC. For this image we generate codebook of size 128 using LBG algorithm and converting them to 8 segmented images are shown in Fig.2(b)-(i). After using Canny's operator the results are displayed in Fig.3(a)-(h). Edge detected images Fig.3(a)-(h) were superimposed on original MRI tumor image Fig.2(a) and displayed as Fig.4(a)-(h) respectively which indicate textural variation as we move from one code-vector to the next. Fig.5(a) shows result for probability using GLCM and equalized probability is displayed in Fig 5(b).Extracted Entropy using GLCM as shown in Fig.5(c) with equalized entropy in Fig.5(d). Fig.6(a) shows superimposed edge map on original image for equalized Entropy using GLCM, Fig.6(b) displays similarly constructed image using watershed algorithm and Fig.6(c) indicates result for superimposed image for first code-vector amongst 8 code-vectors using LBG algorithm.

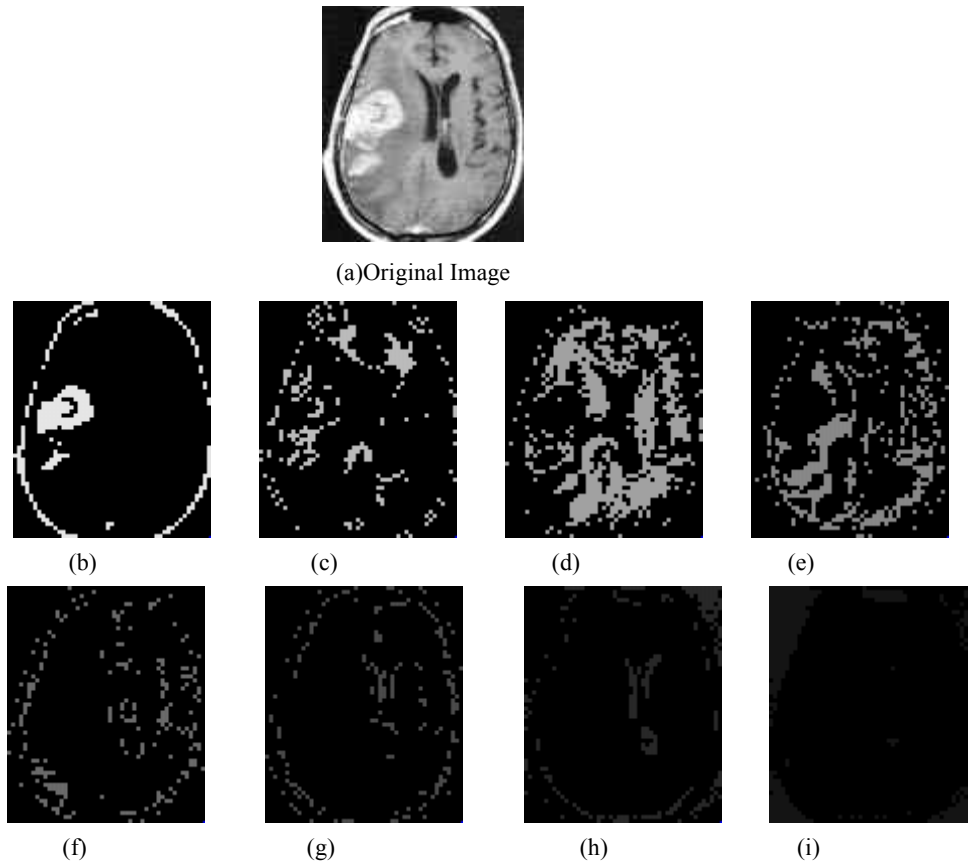

**Fig. 2: (a) Original brain tumor image,(b) Image for first code-vector, (c)-(i) Image for second -eighth code-vector,**

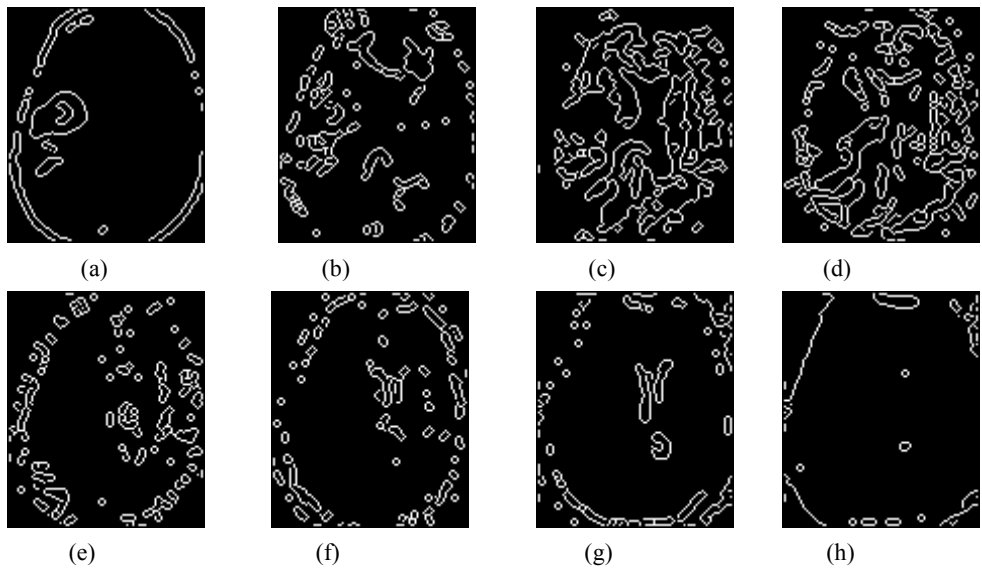

**Fig.3: (a)-(h) Edge detected images for Fig.2(b)-(i) respectively.**

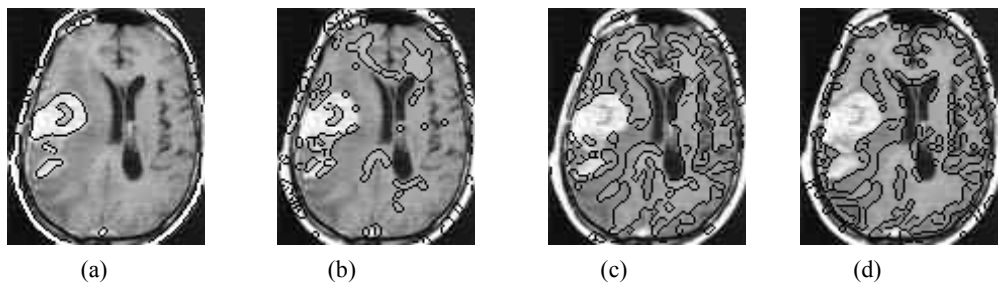

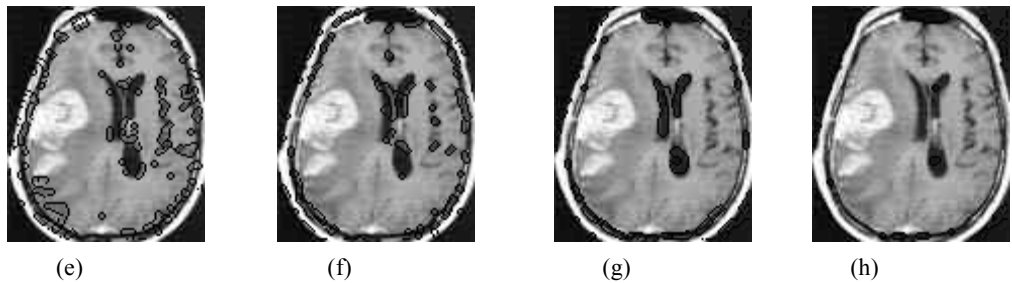

(e)   (f)   (g)   (h)

**Fig.4: (a)-(h) Superimposed images of Fig.3 (a)-(h) respectively on original image Fig.2(a)**

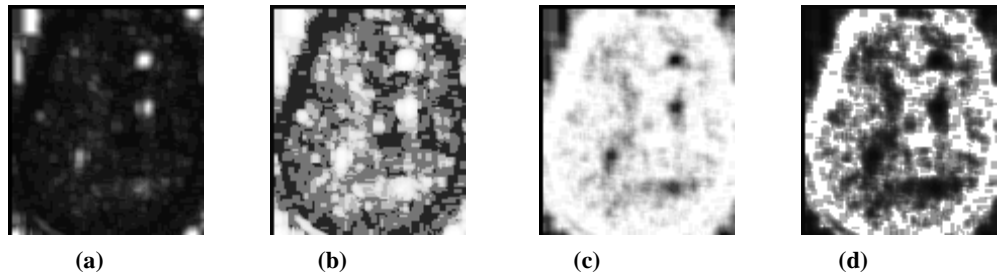

(a)   (b)   (c)   (d)

**Fig. 5 : (a) Extracted Probability using GLCM of Fig 2a, (b) Equalized Probability for Fig.5a,**

**(c) Entropy using GLCM for Fig.2a,(d) Equalized Entropy for Fig.5c.**

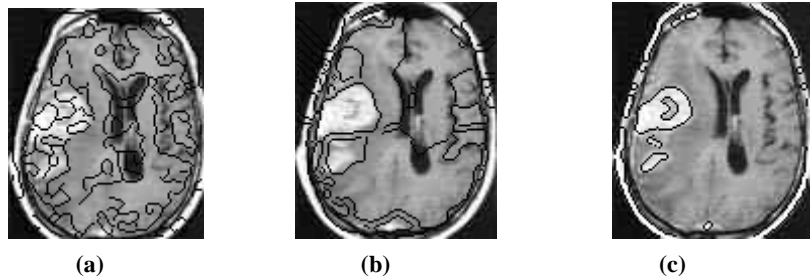

(a)   (b)   (c)

**Fig.6: (a) Segmented image for entropy using GLCM, (b) Segmented image using watershed algorithm,**

**(c) Segmented image using proposed algorithm.**

## 4. CONCLUSION

Here we used Gray Level Co-occurrence Matrix, watershed algorithm and LBG algorithm for tumor detection and demarcation for MRI images. Initially a codebook of size 128 was generated for these images. These code vectors were further clustered in 8 clusters using same LBG algorithm. These 8 images were displayed as a results in Fig 2(b)-(i). From results (Fig.6) it is observed that GLCM, watershed gives over segmentation while LBG shows far better results for the same. This approach does not lead to over segmentation or under segmentation with less complexity.

## 5. ACKNOWLEDGEMENTS

Authors sincerely would like to thank Dr. Manisha Mundhe for identifying tumor and approving the results.

## AUTHOR BIOGRAPHIES

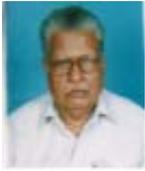

**Dr. H. B. Kekre** has received B.E. (Hons.) in Telecomm. Engg. from Jabalpur University in 1958, M.Tech (Industrial Electronics) from IIT Bombay in 1960, M.S.Engg. (Electrical Engg.) from University of Ottawa in 1965 and Ph.D. (System Identification) from IIT Bombay in 1970. He has worked Over 35 years as Faculty of Electrical Engineering and then HOD Computer Science and Engg. at IIT Bombay. For last 13 years worked as a Professor in Department of Computer Engg. at Thadomal Shahani Engineering College, Mumbai. He is currently Senior Professor working with Mukesh Patel School of Technology Management and Engineering, SVKM's NMIMS University, Vile Parle(w), Mumbai, INDIA. His areas of interest are Digital Signal processing , Image Processing and computer networks. He has more than 250 papers in National / International Conferences / Journals to his credit. Recently six students working under his guidance have received best paper awards. Currently he is guiding ten Ph.D. students.

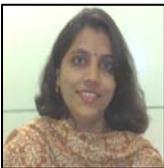

**Tanuja K. Sarode** has Received M.E.(Computer Engineering) degree from Mumbai University in 2004, currently Pursuing Ph.D. from Mukesh Patel School of Technology, Management and Engg., SVKM's NMIMS University, Vile-Parle (W), Mumbai, INDIA. She has more than 10 years of experience in teaching. Currently working as Assistant Professor in Dept. of Computer Engineering at Thadomal Shahani Engineering College, Mumbai. She is life member of IETE, member of International Association of Engineers (IAENG) and International Association of Computer Science and Information Technology (IACSIT), Singapore. Her areas of interest are Image Processing, Signal Processing and Computer Graphics. She has 35 papers in National /International Conferences/journal to her credit.

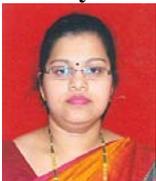

**Ms. Saylee M. Gharge** has received M.E.(Electronics and telecomm.) degree from Mumbai University in 2007, currently Pursuing Ph.D. from Mukesh Patel School of Technology, Management and Engineering, NMIMS University, Vile-Parle (W), Mumbai. She has more than 9 years of experience in teaching. Currently working as a lecturer in department of electronics and telecommunication in Vivekanand Institute of Technology, Mumbai . Her areas of interest are Image Processing, Signal Processing. She has 19 papers in National /International Conferences/journal to her credit